\documentclass[sigconf]{acmart}
\usepackage{algorithm}
\usepackage{algorithmic}
\usepackage{array}
\usepackage{stfloats}
\usepackage{float}
\AtBeginDocument{%
  }

\copyrightyear{2025}
\acmYear{2025}
\setcopyright{cc}
\setcctype{by}
\acmConference[SCF '25]{ACM Symposium on Computational Fabrication}{November 20--21, 2025}{Cambridge, MA, USA}
\acmBooktitle{ACM Symposium on Computational Fabrication (SCF '25), November 20--21, 2025, Cambridge, MA, USA}
\acmDOI{10.1145/3745778.3766670}
\acmISBN{979-8-4007-2034-5/2025/11}

\begin{document}

\title{Speech to Reality: On-Demand Production using Natural Language, 3D Generative AI, and Discrete Robotic Assembly}



\author{Alexander Htet Kyaw}
\orcid{0000-0002-6020-4529}
\affiliation{%
\department{Computer Science and Artificial Intelligence Lab}
  \institution{Massachusetts Institute of Technology}
  \city{Cambridge}
  \state{Massachusetts}
  \country{United States}}
\email{alexkyaw@mit.edu}

\author{Miana Smith}
\orcid{0000-0003-2963-8086}
\affiliation{%
\department{Center for Bits and Atoms}
  \institution{Massachusetts Institute of Technology}
  \city{Cambridge}
  \state{Massachusetts}
  \country{United States}}
\email{miana@mit.edu}

\author{Se Hwan Jeon}
\orcid{0000-0002-2791-7850}
\affiliation{%
\department{Department of Mechanical Engineering}
  \institution{Massachusetts Institute of Technology}
  \city{Cambridge}
  \state{Massachusetts}
  \country{United States}}
\email{sehwan@mit.edu}

\author{Neil Gershenfeld}
\orcid{0000-0001-8470-5777}
\affiliation{%
\department{Center for Bits and Atoms}
  \institution{Massachusetts Institute of Technology}
  \city{Cambridge}
  \state{Massachusetts}
  \country{United States}}
\email{gersh@cba.mit.edu}

\renewcommand{\shortauthors}{Kyaw et al.}

\begin{abstract}
  We present a system that transforms speech into physical objects using 3D generative AI and discrete robotic assembly. By leveraging natural language, the system makes design and manufacturing more accessible to people without expertise in 3D modeling or robotic programming. While generative AI models can produce a wide range of 3D meshes, AI-generated meshes are not directly suitable for robotic assembly or account for fabrication constraints. To address this, we contribute a workflow that integrates natural language, 3D generative AI, geometric processing, and discrete robotic assembly. The system discretizes the AI-generated geometry and modifies it to meet fabrication constraints such as component count, overhangs, and connectivity to ensure feasible physical assembly. The results are demonstrated through the assembly of various objects, ranging from chairs to shelves, which are prompted via speech and realized within 5 minutes using a robotic arm.
\end{abstract}


\begin{teaserfigure}
  \includegraphics[width=1\textwidth]{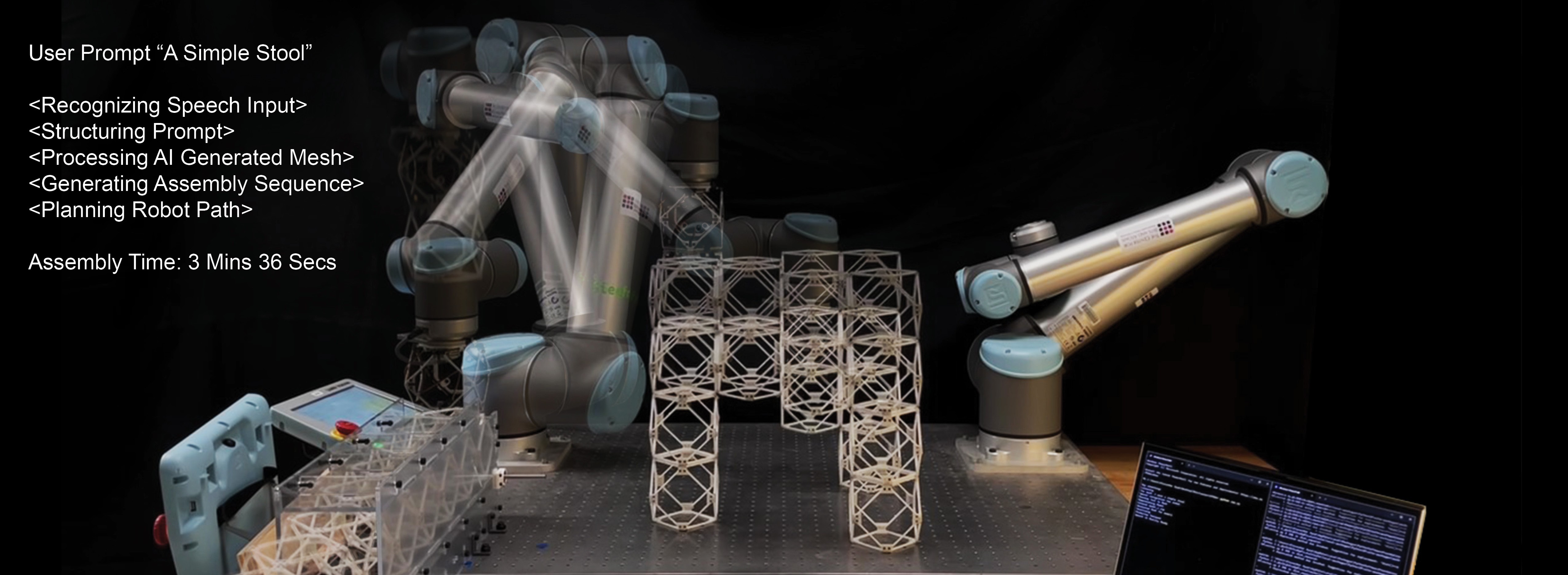}
  \caption{The speech to reality system turning a prompt into a physical object via 3D generative AI and robotic assembly.}
  \Description{blank}
  \label{fig:teaser}
\end{teaserfigure}


\maketitle

\section{Introduction}
Recent advances in 3D generative AI are changing the future of design and manufacturing by allowing the rapid creation of 3D digital assets. Tools such as Get3D \cite{gao_get3d_2022}, Shap-E \cite{jun_shap-e_2023}, AssetGen \cite{siddiqui_meta_2024} can transform text prompts into 3D shapes in seconds. With decreasing generation times and computational costs, the potential for instant, natural language-driven design and manufacturing is becoming increasingly feasible. The ability to create physical objects through speech input could enable people to create objects on demand by simply articulating their needs \cite{ballagas_exploring_2019}. However, translating these digital creations from 3D generative AI or text-to-3D models into physical objects remains a challenge due to fabrication constraints and manufacturing time \cite{abdelaal_ai_2024}. 

To address these constraints, this paper presents an automated system that transforms speech into physical objects through generative AI and discrete robotic assembly (Fig. \ref{fig:teaser}). Generative AI models can generate a wide variety of geometries, requiring a workflow that can adapt to user prompts, geometric variability, and physical constraints. Our approach utilizes a Large Language Model to process natural language into structured input for generative AI, a discretization method to convert AI-generated meshes into component-level representations suitable for robotic assembly, and geometric processing for fabrication constraints. 

We identified key fabrication constraints and developed geometric processing steps to automatically modify AI-generated meshes for assembly feasibility. Assembly feasibility is defined as the condition under which an AI-generated design (a) fits within the robot’s workspace, (b) uses no more components than are available in inventory, (c) remains within the allowable limits for unsupported overhangs and vertical stacks, and (d) maintains face connectivity from the ground to all components. 

Additionally, Generative AI models can rapidly produce 3D models, making them well-suited for iterative design workflows. While generative models can produce a variety of digital objects in seconds, physically realizing these designs can be slow, resource-intensive, and often unsustainable at scale \cite{kyaw_making_2025} . Rather than aiming to replace traditional manufacturing methods, we argue that there is a need to explore new pipelines for creative co-creation between humans, AI, and robotic systems. Through this paper, we present three key contributions:

\begin{itemize}
    \item \textbf An integrated pipeline that connects a text-to-3D generative AI model with discrete robotic assembly.
    \item \textbf Geometric processing methods that enable the assembly of AI-generated objects by enforcing fabrication constraints.
    \item \textbf A demonstration of how the \textit{Speech-to-Reality} framework aligns with the speed and generative capacity of 3D AI models to support on-demand and sustainable production (Fig.~\ref{fig:teaser}).
\end{itemize}

\section{Related Work}

\subsection{3D Generative AI for Physical Objects}

3D generative AI models such as  DreamFusion \cite{poole_dreamfusion_2022-1}, Neuralangelo \cite{li_neuralangelo_2023}, and LATTE3D \cite{xie_latte3d_2024} are capable of generating meshes in just seconds, lowering the barrier of entry to 3D content creation.


Despite these advancements in 3D generative AI, there remains limited work on translating the outputs from AI generated output into physical objects \cite{li_generative_2024}. Previous efforts using generative AI to create physical objects have focused mainly on 3D printing \cite{danry_organs_2023}. For example, prior work includes a framework that integrates text and sketch input to enhance the manufacturability of AI-generated designs for 3D printing \cite{edwards_sketch2prototype_2024}, a workflow that uses generative design to create parts compatible with 3D printing or CNC machining constraints.  \cite{mcclelland_generative_2022}, and a system that applies generative AI to stylize existing 3D models based on functionality for 3D printing \cite{faruqi_style2fab_2023}. These approaches predominantly emphasize the use of Generative AI for 3D printing or CNC machining of small objects or parts. Although generative AI can create 3D models of any scale in seconds, conventional digital fabrication can take hours or days, depending on the size of the object, leading to a disconnect between AI capabilities and physical production. 

\subsection{Natural Language for 3D Geometry}

Recent studies have explored how Large Language Models (LLMs) can be embedded within Computer Aided Design (CAD) software to simplify design tasks \cite{zhang_large_2025}. Previous projects have demonstrated using text and images to modify CAD models. \cite{daareyni_generative_2025}. Plugins such as CADgpt for Rhino3D demonstrate that natural language can drive parametric modeling and shape manipulation, potentially reducing the learning curve for new users \cite{kapsalis_cadgpt_2024}. While these systems shift design into natural language, they still rely on manual downstream steps for physical fabrication: users must export models, slice for printing/CAM, and assemble parts themselves. Makatura et al. showcased the potential of ChatGPT in design and manufacturing by using it to generate the design of a laser-cut shelf \cite{makatura_how_2023}. While AI tools like GPT and Text-to-Mesh models can sometimes create designs that can be digitally fabricated, manual assembly is still required. Therefore, this paper introduces a fully automated approach that connects natural language, 3D generative AI, and robotic assembly to integrate the entire production process. 

\subsection{Automated Robotic Assembly Workflows}

Prior research has focused on automating various aspects of the robotic assembly process. Tian et al. present a physics-based method for assembly sequence planning using graph neural networks \cite{tian_asap_2024}. Gandia et al. present a path planning workflow that adjusts path planning parameters according to the assembly geometry \cite{gandia_towards_2019}. Macaluso et al. demonstrated the use of ChatGPT for robotic programming by decomposing complex tasks into sub-tasks \cite{macaluso_toward_2024}. 
These studies have mainly explored automated robotic assembly workflows with human-generated designs or conventional CAD software. However, integrating the outputs of 3D generative AI model may require different considerations in fabrication constraints, time, and sustainability. Generative AI can rapidly create digital assemblies, but producing physical ones can still be time-consuming. Each new design requires a new set of components to be fabricated before assembly. While it may be hard to replace conventional CAD workflows, this paper proposes integrating generative AI with discrete modular robotic assembly.


\subsection{Modular and Discrete Systems}

Modular systems highlight how prefabricated components can be reused without the need to manufacture new parts for each assembly. 
Previous studies on Voxel-based systems have demonstrated sustainable frameworks where unit cells, components, or voxels are designed for ease of connection and reassembly \cite{jenett_materialrobot_2019}, \cite{abdel-rahman_self-replicating_2022}, \cite{gregg_ultralight_2024}, \cite{smith_self-reconfigurable_2024}, \cite{smith_voxel_2025}. 
However, prior research relied on manually designed predefined structures. In contrast, this paper introduces a novel framework where the outputs from a 3D generative AI model determine the assembly. To support this, the Speech-to-Reality system presents an automated pipeline that transforms user inputs into feasible voxel-based assemblies while accounting for the variability of AI generated meshes, user prompts, and fabrication constraints. 

 
\section{Methods}

\begin{figure*}[b]
  \centering
  \includegraphics[width=\textwidth]{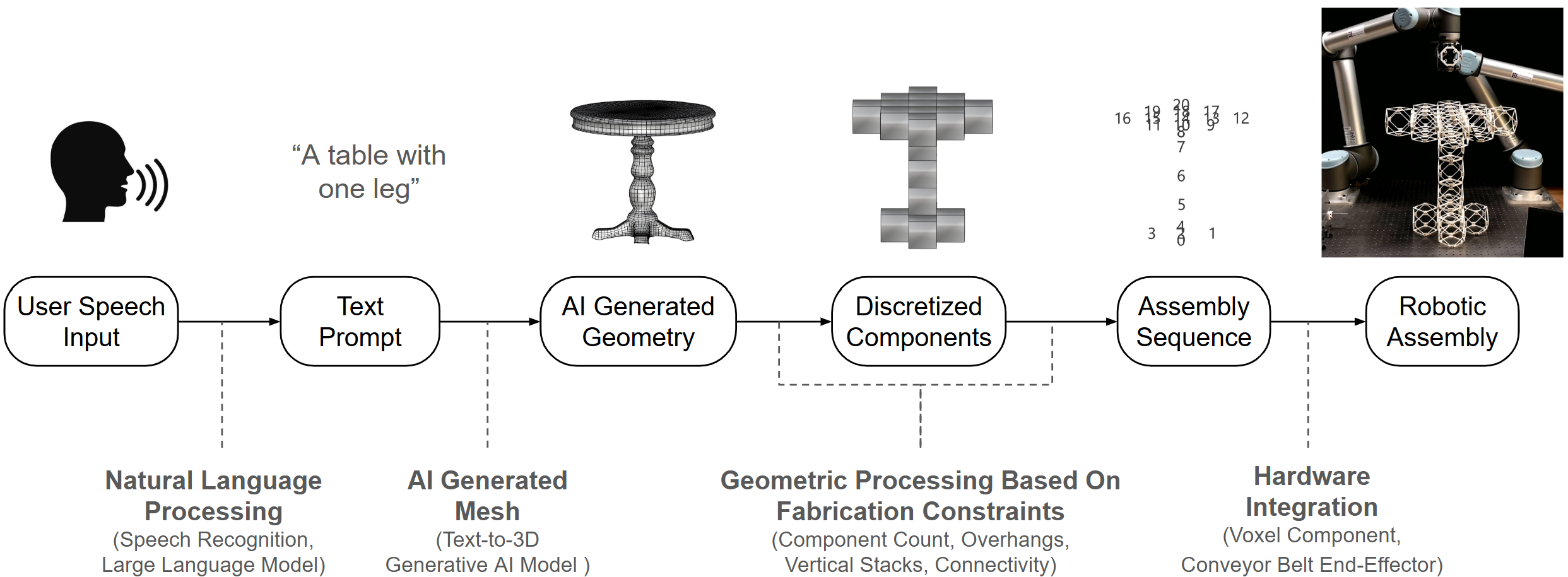}
  \caption{Data pipeline and software components of the speech-to-reality system.}
  \Description{Diagram showing the data pipeline and software components of the speech-to-reality system.}
  \label{fig:pipeline}
\end{figure*}

Speech to Reality is an automated system that integrates: (a) natural language processing, (b) 3D 3D Generative AI , (c) geometric processing for fabrication constraints, and (d) hardware integration for robotic assembly (Fig.~\ref{fig:pipeline}). A key contribution of this work is identifying and integrating all the necessary components to go from 3D Generative AI to discrete robotic assembly, in a way that addresses fabrication constraints, enables user-driven on-demand production, and supports sustainable creation. Since the pipeline is modular, each module can be swapped for alternative implementations. While we present a specific implementation, the overall workflow is extendable to other tools.

\subsection{Natural Language Processing to Structured Input for Generative AI }

To process speech into structured input, the system first converts spoken language into text. It then uses a LLM, GPT-4 Turbo \cite{openai_openai_2025},  to analyze the transcription and identify the object the user wants to assemble. The LLM is tasked with distinguishing between actionable commands involving physical objects and non-physical concepts unsuitable for 3D generative AI by using a guided prompt. The guided prompt was structured to assist the language model interpret various user inputs effectively. If the prompt identifies a physical object, the system extracts and returns it as a response. If no object is detected, the response returns "false," and the system would ask the user to restate their command to request a physical object. 
The prompt we used is: "Your task is to analyze the given text and determine whether it refers to a physical object or shape that is not an abstract idea. If it refers to something physical, return the relevant phrase that describes it; otherwise, respond with 'false." To ensure the correct format, the LLM is also given the two examples listed below.

• Object Request: "I need a shelf " → Response: "shelf"

• Non-Object Request: "Knowledge" → Response: "false" 

\subsection{AI-Generated Mesh to Component Discretization for Robotic Assembly}

The LLM output is used as a text prompt for the generative AI model to create a mesh. However, the outputs from 3D generative AI models are typically in the form of meshes or point clouds, which are not directly suitable for robotic assembly. To address this, we developed a component discretization algorithm that converts an AI-generated mesh into a component-level representation suitable for robotic assembly. 

The script starts by post-processing the AI-generated mesh by unifying normals, welding vertices, and eliminating geometric inconsistencies. The bounding box of the mesh is then calculated along the x, y, and z axes to determine its overall dimensions. To ensure that the object can be physically assembled within the workspace, the bounding box and the mesh is uniformly scaled to fit the assembly space. This also verifies that all the component positions fall within the reachable workspace of the robotic arm, ensuring that each part can be placed without violating the arm’s kinematic constraints. In the discretization process, the bounding box encapsulating the scaled AI generated mesh is divided into a 3D grid by generating planes along the x, y, and z axes. These planes are spaced at regular intervals based on the size of the individual components. In our experiments, we use modular components measuring 10 cm × 10 cm × 10 cm. After the bounding box is divided into a 3D grid based on the size of the components, a Boolean intersection algorithm checks whether each grid cell contains a part of the mesh. (Fig. \ref{fig:three}) If any portion of the mesh is inside a cell, it is marked as \textbf{true}. Otherwise, it is marked as \textbf{false}. The script then automatically assigns an assembly component to every grid cell marked true, generating a component-level representation of an AI-generated 3D mesh that can be robotically assembled. See Algorithm \ref{alg:1} for more details. 

In this study, we used Meshy.ai as the 3D generative AI model \cite{meshy_meshy_2025}, which produces meshes aligned to a Cartesian grid. This ensures a consistent orientation, with the Z-axis of each object always upright. Aligning output meshes to a Cartesian grid is important because the orientation directly affects the results of the discretization process. 



\begin{algorithm}
\caption{Component Discretization for Robotic Assembly}
\label{alg:1}
\begin{algorithmic}
\REQUIRE AIGeneratedMesh, ComponentSize, AssemblySpace
\ENSURE AssemblyGeometry is component-level representation of AIGeneratedMesh

\STATE \textbf{function} \textsc{ComputeBoundingBox}(AIGeneratedMesh): Compute bounding box along $x, y, z$ axes of a geometry
\STATE \textbf{function} \textsc{ScaleToFit}(AIGeneratedMesh, BoundingBox, AssemblySpace): Uniformly scale geometry to fit the AssemblySpace
\STATE \textbf{function} \textsc{Discretize}(BoundingBox, ComponentSize): Generate 3D grid over BoundingBox with spacing from ComponentSize
\STATE \textbf{function} \textsc{Intersect}(GridCell, AIGeneratedMesh): Returns \textbf{true} if GridCell intersects AIGeneratedMesh

\STATE BoundingBox = \textsc{ComputeBoundingBox}(AIGeneratedMesh)
\STATE AIGeneratedMesh = \textsc{ScaleToFit}(AIGeneratedMesh, BoundingBox, AssemblySpace)

\STATE Grid = \textsc{Discretize}(BoundingBox, ComponentSize)
\STATE AssemblyGeometry = $\emptyset$

\FOR{each GridCell in Grid}
    \IF{\textsc{Intersect}(GridCell, AIGeneratedMesh) == \textbf{true}}
        \STATE Add component at GridCell to AssemblyGeometry
    \ENDIF
\ENDFOR

\RETURN AssemblyGeometry
\end{algorithmic}
\end{algorithm}

\subsection{Geometric Processing of AI-Generated Designs for Fabrication Constraints}

Assembly feasibility depends on whether an AI-generated geometry can be physically constructed using the selected components and robotic system. From an AI-generated mesh, the previous step produced a component-level assembly geometry that provides the assembly coordinates. The system applies geometric processing techniques to analyze and modify these coordinates to satisfy fabrication constraints. In this research, we identify key fabrication constraints that a pipeline like Speech-to-Reality must handle to account for the variability of AI-generated assemblies. These constraints include the number of components, overhanging elements, connectivity between parts, and the reachability of the robotic arm.

\subsubsection{Component Count}

AI-generated designs can be decomposed into assemblies that require varying numbers of components. However, in real-world settings, the number of physical components available may be limited, which introduces a constraint on what can be built. In our setup, we only have 40 physical components for assembly. To address this, the system counts the total number of components in the assembly and checks if it exceeds the number of available physical components. If the count doesn’t exceed the number of available components, the assembly passes the component count check. Otherwise, it fails. The system has an automated failure handling mechanism that makes modifications to the assembly geometry. When the component count exceeds the available number of physical components, the system iteratively applies a uniform scale reduction equivalent to the size of a single component, then reruns the discretization process. This process continues until the total component count falls within the allowable limit. See Algorithm \ref{alg:2} for more details. 

\begin{algorithm}
\caption{Component-Count Check with Auto-Rescaling}
\label{alg:2}
\begin{algorithmic}
\REQUIRE AssemblyGeometry,\quad \textit{MaxComponents} 
\ENSURE Modified geometry produces no more than \textit{MaxComponents} components

\STATE \textbf{function} \textsc{Discretize}(\textit{geometry}): Returns grid cells occupied by geometry based on component size
\STATE \textbf{function} \textsc{Scale}(\textit{geometry}): Uniformly scales down the assembly geometry by one component size

\STATE $AssemblyCoordinates = \textsc{Discretize}(AssemblyGeometry)$
\STATE $AssemblyCount = |AssemblyCoordinates|$

\WHILE{$AssemblyCount >$ \textit{MaxComponents}}
    \STATE $AssemblyGeometry = \textsc{Scale}(AssemblyGeometry)$
    \STATE $AssemblyCoordinates = \textsc{Discretize}(AssemblyGeometry)$
    \STATE $AssemblyCount = |AssemblyCoordinates|$
\ENDWHILE

\RETURN $AssemblyGeometry$
\end{algorithmic}
\end{algorithm}

\subsubsection{Overhang Detection and Vertical Stacks}
Although AI-generated assemblies can take on various shapes in the digital world, certain geometric configurations, such as cantilevers or tall, unsupported stacks, may lead to failure during physical fabrication. To evaluate real-world constraints, we conducted physical empirical tests using our modular components and found that cantilevers extending beyond three unsupported elements and vertical stacks taller than four components without lateral support tend to be unstable. This number may vary for other types of components and robot setups.

Based on this finding, we implemented an overhang detection algorithm. The system performs an overhang check by identifying any components that lack vertical support and extend directionally beyond the overhang limit. If a cantilever exceeds the allowable limit, the system iteratively rescales the geometry along the axis of the overhang by one unit, repeating this process until all overhang violations are resolved. 
A similar procedure is applied to vertical stability: the system detects columns that are vertically stacked without sufficient lateral support. If any such stack exceeds the predefined height threshold, in our case this number is four, the geometry is scaled down along the vertical axis until the violation is resolved. See Algorithm \ref{alg:3} for implementation details.

\begin{algorithm}
\caption{Overhang Check with Dynamic Rescaling}
\label{alg:3}
\begin{algorithmic}
\REQUIRE AssemblyGeometry, OverhangLimit, VerticalLimit
\ENSURE AssemblyGeometry is within the OverhangLimit 

\STATE \textbf{function} \textsc{Discretize}(\textit{geometry}): Returns grid cells occupied by geometry based on component size
\STATE \textbf{function} \textsc{Scale1D}(\textit{geometry}, \textit{axis}): Uniformly scales down geometry along \textit{axis} by one component size

\STATE $AssemblyCoordinates \gets$ \textsc{Discretize}$(AssemblyGeometry)$

\FOR{each cell $(x, y, z)$ with no support at $(x, y, z{-}1)$}
    \STATE $d \gets$ number of unsupported cantilever components
    \STATE $\hat{d}_{\text{axis}} \gets$ direction of overhang (e.g., $\hat{x}$ or $\hat{y}$)
    \WHILE{$d >$ OverhangLimit}
        \STATE $AssemblyGeometry \gets$ \textsc{Scale1D}$(Geometry, \hat{d}_{\text{axis}})$
        \STATE $AssemblyCoordinates \gets$ \textsc{Discretize}$(Geometry)$
        \STATE Recalculate $d$ and $\hat{d}_{\text{axis}}$
    \ENDWHILE
\ENDFOR

\FOR{each $(x, y, z)$ with no lateral support at $(x \pm 1, y \pm 1, z)$}
    \STATE $h \gets$ number of consecutive unsupported vertical stacks
    \WHILE{$h >$ VerticalLimit}
        \STATE $AssemblyGeometry \gets$ \textsc{Scale1D}$(Geometry, \hat{z})$
        \STATE $AssemblyCoordinates \gets$ \textsc{Discretize}$(Geometry)$
        \STATE Recalculate $h$
    \ENDWHILE
\ENDFOR

\RETURN $AssemblyGeometry$
\end{algorithmic}
\end{algorithm}

This directional rescaling mechanism ensures that the final assembly remains within the fabrication constraints without uniformly shrinking the entire geometry. Instead of simply removing overhanging or unstable components, which could leave structural gaps and compromise the intended design, the scaling approach adapts the geometry incrementally while preserving its overall form. 
See Algorithm 3 for implementation details.

\subsubsection{Component Connectivity and Assembly Sequence}In an AI-generated assembly, the components are not sorted in any specific order. However, in discrete robotic assembly, a component can only be placed if it is connected to either the ground or an already placed component. Additionally, the assembly sequence needs to be sorted to prevent the robotic arm from colliding with previously assembled components.

\begin{figure}[!]
    \centering
    \includegraphics[width=0.99\linewidth]{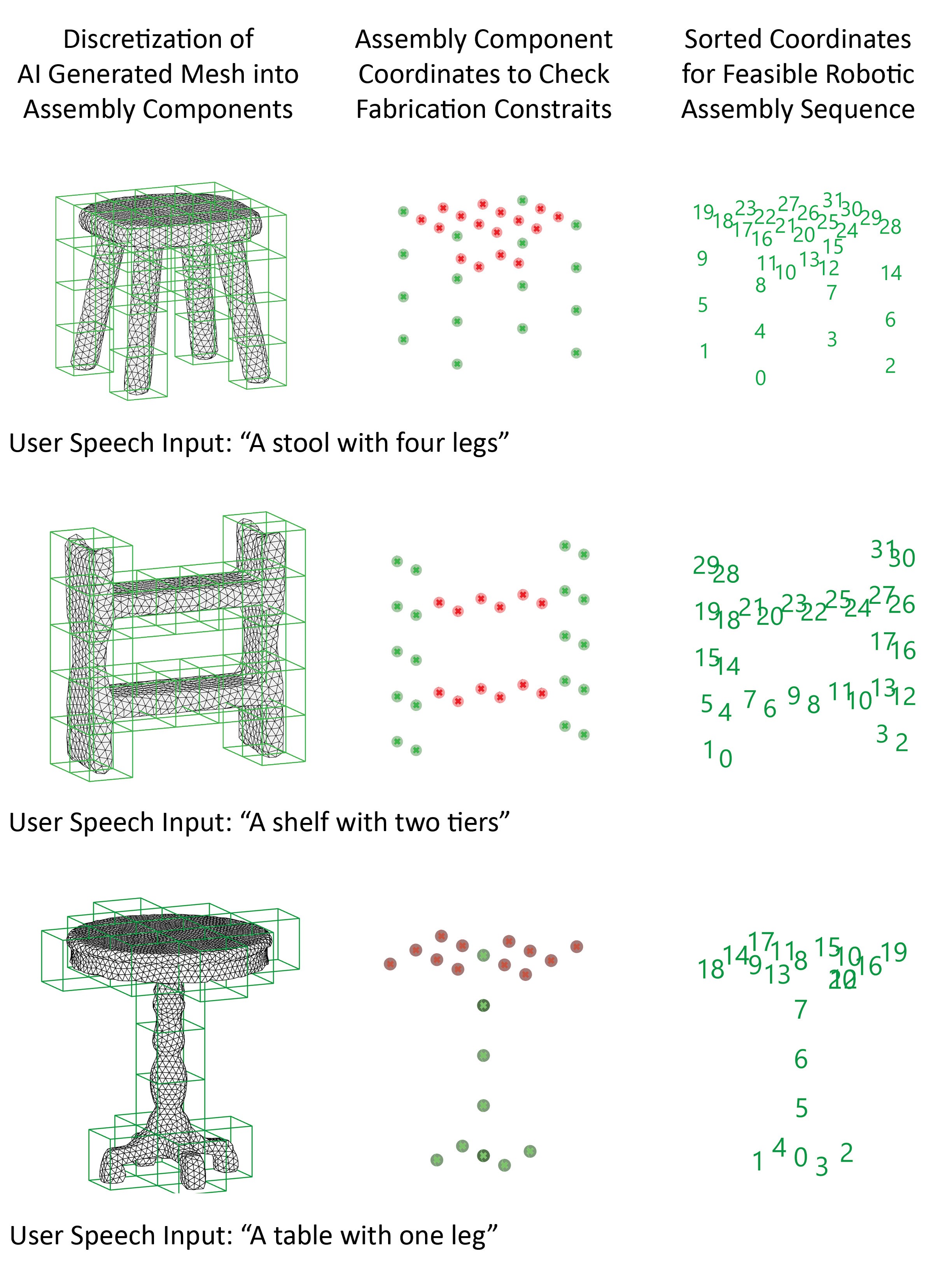}
    \Description{A sequence showing AI generated mesh, discretized components, assembly coordinates, and sorted assembly sequence.}
    \caption{From AI generated mesh to discretized components, to assembly coordinates, to sorted assembly sequence.}
    \label{fig:three}
\end{figure}

To address this, the sequence is first sorted by z-values, enabling the robot to construct the assembly layer by layer from the bottom up. Within each layer, the components are further sorted by x-values, followed by y-values. However, this simple sorting method does not fully account for structural connectivity. To ensure that each component shares at least one face with a previously placed component, we implement a connectivity search algorithm. This algorithm prioritizes components based on their proximity to already assembled ones. Specifically, it searches for the components with the shortest distance to a previously placed component. The optimized assembly sequence is saved as a sorted list of coordinates (Fig.~\ref{fig:three}). See Algorithm \ref{alg:4}. To sum up, the geometric processing starts from an AI-generated mesh, to discretized assembly components, to assembly coordinates, and finally to an ordered assembly sequence. 

\begin{algorithm} 
\caption{Connectivity-Aware Assembly Sequence Generation}
\label{alg:4}
\begin{algorithmic}
\REQUIRE AssemblyCoordinates 
\ENSURE Sorted AssemblyCoordinates
\STATE Sort $AssemblyCoordinates$ by increasing $z$, then $x$, then $y$
\STATE $Placed \gets \emptyset$ {Set of already placed components}
\STATE $Sequence \gets [\,]$ {Final ordered sequence}
\WHILE{$AssemblyCoordinates \neq \emptyset$}
    \FOR{each $coord$ in $AssemblyCoordinates$}
        \IF{$coord.z = 0$ \OR $coord$ is adjacent to any $p \in Placed$}
            \STATE Add $coord$ to $Sequence$
            \STATE Add $coord$ to $Placed$
            \STATE Remove $coord$ from $AssemblyCoordinates$
        \ENDIF
    \ENDFOR
    \IF{no component was placed in this iteration}
        \STATE Find $coord \in AssemblyCoordinates$ with minimum distance to any $p \in Placed$
        \STATE Add $coord$ to $Sequence$
        \STATE Add $coord$ to $Placed$
        \STATE Remove $coord$ from $AssemblyCoordinates$
    \ENDIF
\ENDWHILE
\RETURN $Sequence$
\end{algorithmic}
\end{algorithm}

\subsection{Automated Path Planning for Robotic Assembly}

After ensuring that the AI-generated sequence meets fabrication constraints, a path planning procedure is required for robotic assembly. We developed an automated path planning algorithm using the Python-URX library. The algorithm takes in three key parameters. Assembly Coordinates: A list of sorted (x, y, z) coordinates based on the AI-generated object. Source Coordinate: The (x, y, z) position of the component's pick location. Movement Plane: The z-value at which the robot can move safely without colliding with previously assembled components or the conveyor belt. 
 
When the program starts, the robot moves from its resting position to the movement plane. For the pick operation, the robot first moves to the (x, y) position of the source coordinate while remaining in the (z) position of the movement plane. Then it moves to the (z) position of the source coordinate, activates the gripper, picks the component, and returns to the (z) position of the movement plane. 

For the place operation, the robot moves to the (x, y) position of the assembly coordinate while remaining in the (z) position of the movement plane. It then moves to the (z) position of the assembly coordinate, closes the gripper, picks up the component, and returns to the (z) position of the movement plane. These sets of operations are repeated for each component in the sorted assembly sequence. The path planning algorithm repeats this for each coordinate in the sorted assembly sequence until the assembly is complete. See Algorithm \ref{alg:5} for more details.  


\begin{algorithm}
\caption{Path Planning for Automated Assembly}
\label{alg:5}
\begin{algorithmic}
\REQUIRE HomePosition $(x_h, y_h, z_h)$, MovementPlane ($z_m$), AssemblyCoordinates $(x_a, y_a, z_a)$, SourceCoordinate $(x_s, y_s, z_s)$, 

\STATE Move robot to HomePosition $(x_h, y_h, z_h)$
\STATE Move robot to $(x_h, y_h, z_m)$
\FOR{each $(x_a, y_a, z_a)$ in AssemblyCoordinates}
    \STATE Move to $(x_s, y_s, z_m)$
    \STATE Move to SourceCoordinate $(x_s, y_s, z_s)$
    \STATE Gripper Close () Activate gripper to pick component
    \STATE Move to $(x_s, y_s, z_m)$
    \STATE Move to $(x_a, y_a, z_m)$
    \STATE Move to AssemblyCoordinate $(x_a, y_a, z_a)$
    \STATE Gripper Open () Release gripper to place component
    \STATE Move to $(x_a, y_a, z_m)$
\ENDFOR
\STATE Move robot to HomePosition $(x_h, y_h, z_h)$
\end{algorithmic}
\end{algorithm}

\subsection{Prefabricated Components for Discrete Assembly and Disassembly}

\begin{figure}
    \centering
    \includegraphics[width=1\linewidth]{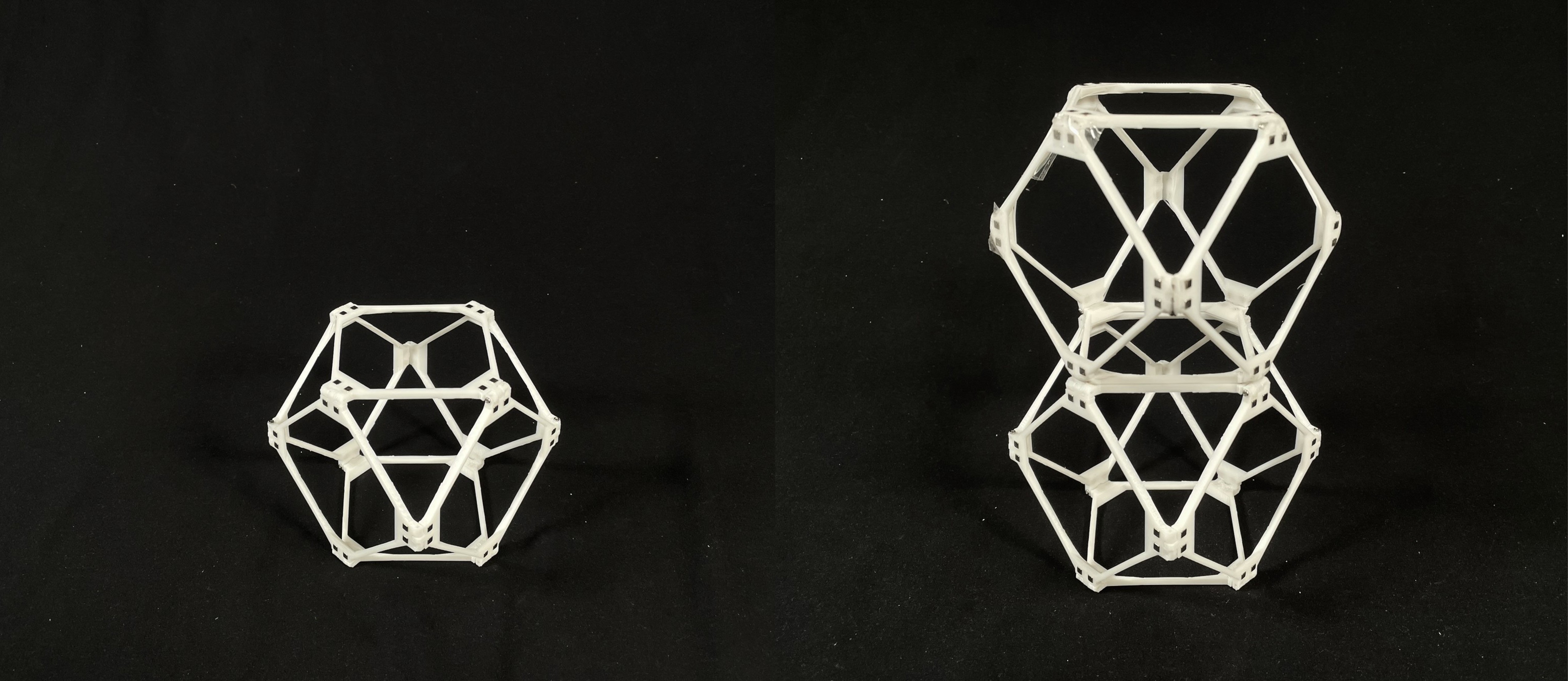}
    \caption{Prefabricated components for robotic assembly}
    \Description{Prefabricated components for robotic assembly shown in the image.}
    \label{fig:voxel}
\end{figure}

The system enables the robotic assembly of modular components that can be assembled and disassembled. Each component is made up of six 3D-printed faces forming a cuboctahedron geometry. Each face is embedded with magnets, ensuring secure attachment between adjacent components while allowing for reversible connections (Fig.~\ref{fig:voxel}). The magnet-based connections allow  fast, tool-free assembly and disassembly. In this research, pre-fabricated components are used to demonstrate the system, not to replace conventional engineering-grade fabrication.

\subsection{Custom End-Effector for Robotic Assembly }

\begin{figure}
    \centering
    \includegraphics[width=1\linewidth]{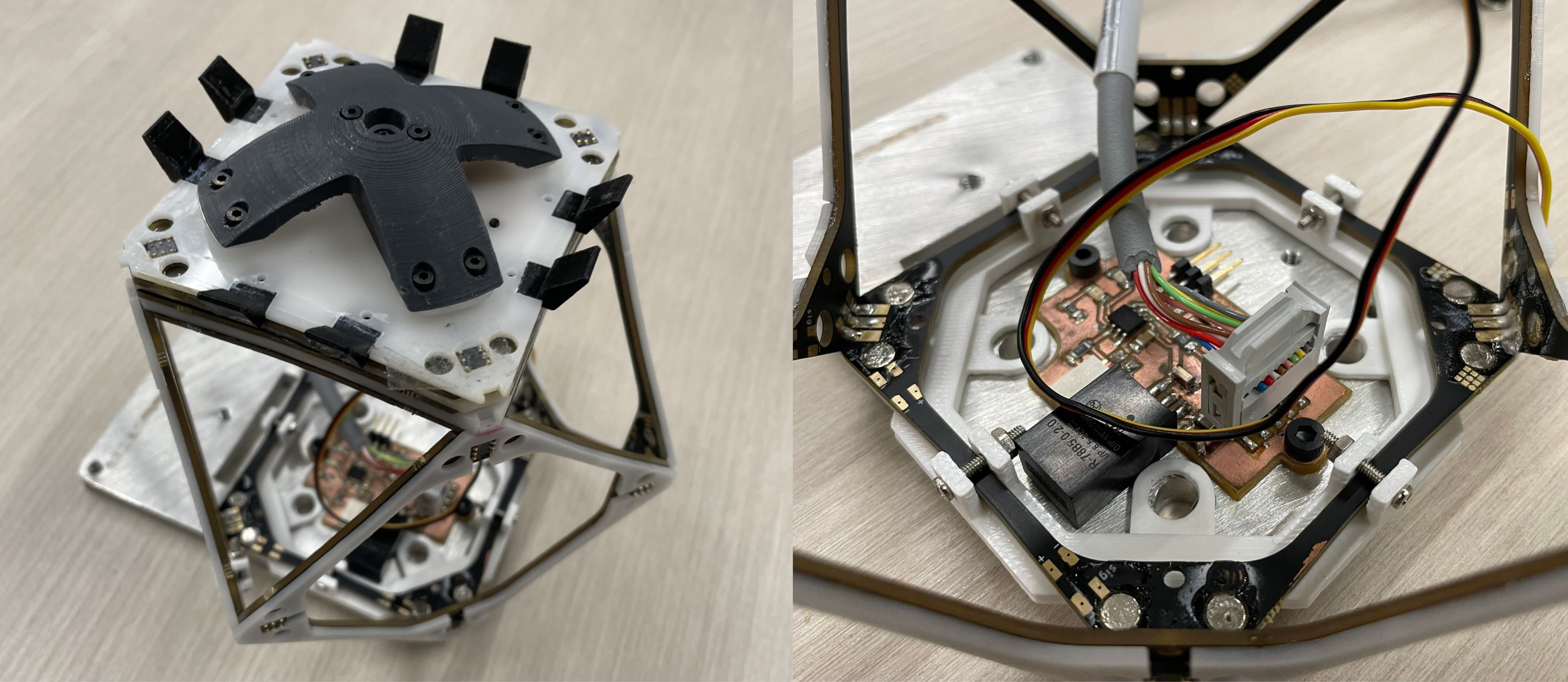}
    \caption{Latch mechanism and indexers of the end effector.}
    \Description{Photo of the latch mechanism and indexers of the end effector.}
    \label{fig:endeffector}
\end{figure}

Since our system repeatedly reuses the same components, we utilize a custom robotic end-effector to ensure consistent assembly. The system employs a 6-axis robotic arm, specifically the Universal Robot UR10 \cite{universal_robots_ur10e_2025}. The gripper end effector is attached to the mounting plate of the robotic arm. Communication between the robot and the gripper is facilitated using the built-in digital I/O pins from the UR10 robotic arm to the ATtiny412 microcontroller. The gripping mechanism follows a design to minimize the use of actively controlled moving parts \cite{jenett_materialrobot_2019}. A single actuator rotates a plus-shaped latch clockwise by 45°, establishing four contact points with the component’s top face (Fig.~\ref{fig:endeffector}). Additionally, reused components may have minor deformations from wear and tear. To address this, the end effector has geometric indexers that ensure precise alignment. These serve as passive self-correcting mechanisms for secure attachment. 

\begin{figure}
    \centering
    \includegraphics[width=1\linewidth]{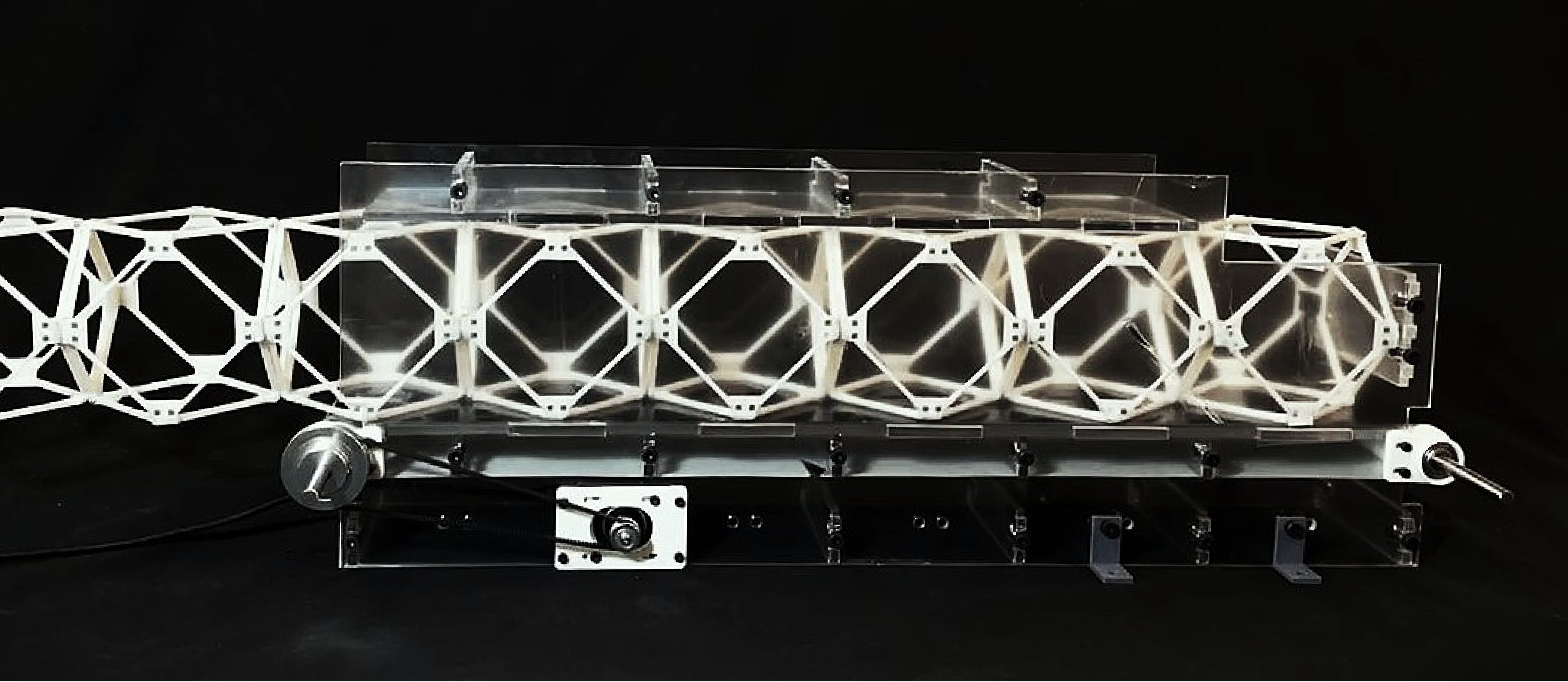}
    \caption{Conveyor belt for automated assembly.}
    \Description{A conveyor belt used for automated assembly.}
    \label{fig:belt}
\end{figure}

\subsection{Conveyor System for Reuse and Sustainable Production}

A custom conveyor belt system is developed to fully automate the robotic assembly process. In our experiments, we used a conveyor belt that can hold a sequence of horizontally connected voxels as feedstock. 
(Fig.~\ref{fig:belt}).
After each assembly, all disassembled components are returned to the conveyor system, making them immediately available for the next object fabrication. This continuous recirculation of components reduces the need for new materials and enhances the efficiency of on-demand fabrication. By enabling the reuse of prefabricated modules across multiple assemblies, the system minimizes material waste and supports a circular model of production. 

\section{Results and Experiments}

\subsection{Natural Language User Input for 3D Generative AI }

\begin{figure*}[b]
    \centering
    \includegraphics[width=\textwidth]{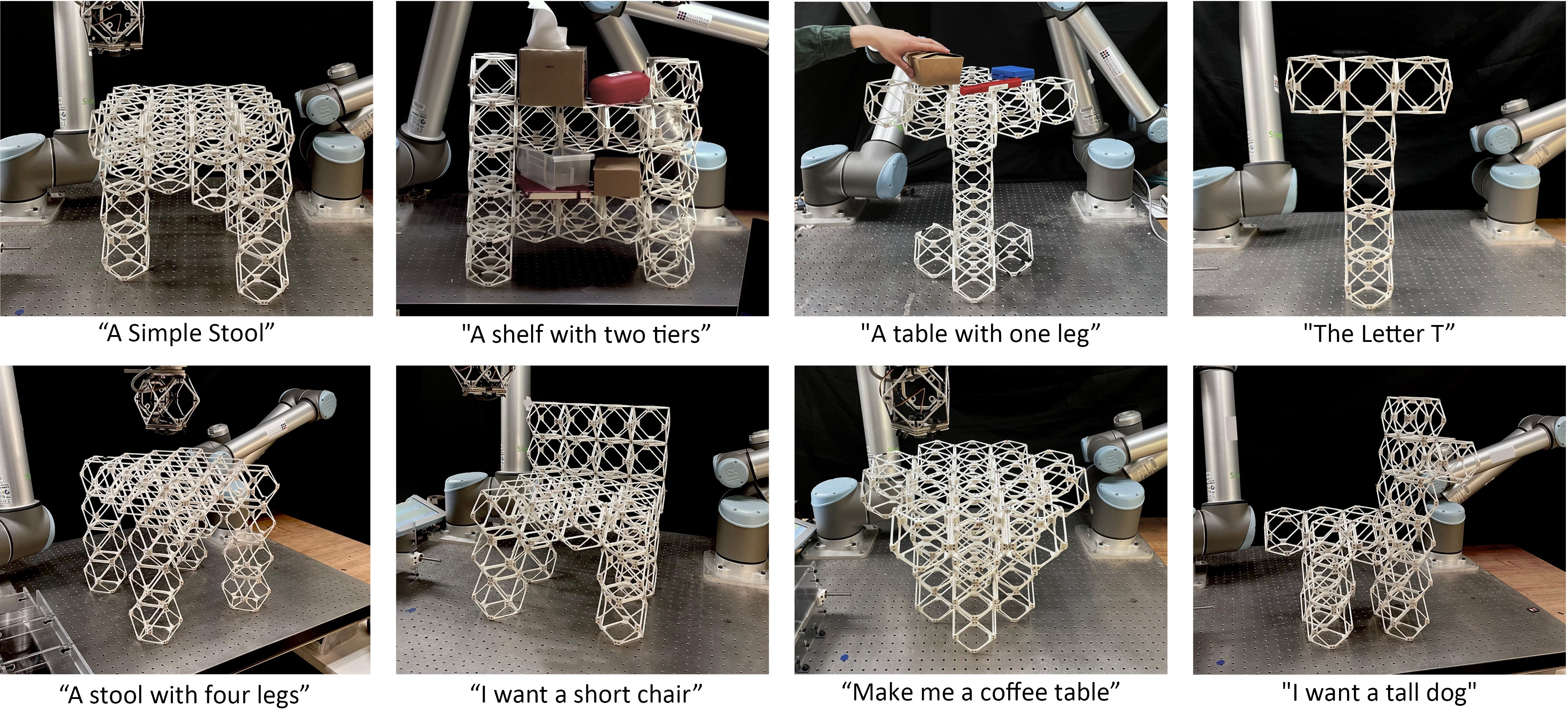}
    \caption{Objects created from the Speech-to-Reality workflow, with components being reused for multiple prompts.}
    \Description{Objects created from the Speech-to-Reality workflow, showing components reused for multiple prompts.}
    \label{fig:all}
\end{figure*}

The system leverages LLMs to interpret user requests and distinguish between abstract concepts and physical objects. For example, it accurately processes commands like "make me a coffee table" as "coffee table" and "I want a simple stool" as "simple stool." It also successfully handles functional specifications such as "a shelf with two tiers," “assemble me a table with one leg” (Fig.~\ref{fig:all}). or "a stool with four legs" (Fig.~\ref{fig:all}). In addition to physical requests, the system effectively identifies abstract prompts, such as "create beauty" or "I need something to hold memories," and correctly labels them as "false." However, the system struggles when abstract concepts are paired with physical requests. For instance, with the input "I need a box to hold memories," it correctly filters out the abstract portion, but still outputs "box." Optimizing the structure of the prompt could improve its ability to handle more nuanced inputs.

\subsection{Fabrication Constraints and Feasibility of AI-Generated Objects via Robotic Assembly }

The system demonstrates various geometric processing methods to ensure the feasibility of AI-generated objects through discrete robotic assembly. The experiments demonstrate that the system can successfully assemble a diverse range of object requests, from functional items like stools, chairs, tables, and shelves to unconventional objects, such as a dog or the letter T (Fig.~\ref{fig:all}). 

To evaluate the contribution of each geometry processing constraint, we performed an ablation study across four user prompts: stool, shelf, letter~T, and table. In the ablation study, we evaluated the contribution of each geometry processing constraint to the overall robustness of the complete pipeline. This is done by comparing the full pipeline against four ablated versions, each missing one of the following modules: (1) component inventory check, (2) overhang check, (3) vertical stack check, and (4) connectivity-aware assembly sequencing. The baseline for all ablation variants begins with the AI-generated mesh, which is uniformly scaled to fit within the assembly workspace (60 cm × 50 cm × 60 cm) and discretized at a 10 cm resolution. 

Table~\ref{tab:validation-results} summarizes the results of the ablation study, where individual geometric processing modules were selectively disabled to evaluate their impact. The findings show that component count is especially critical for objects like the stool and shelf, which occupy a large volume in the assembly space. Overhang detection is important for objects with long cantilevers, such as the shelf. Vertical stacking becomes essential for tall objects like the letter "T". Connectivity-aware sequencing is crucial for objects with branching overhangs, including the stool, shelf, and table. The study reveals conditions under which the system may fail, showing that the importance of each geometry processing step depends on the object's shape. 

The complete pipeline assembled all objects successfully, showing the value of combining multiple constraint checks (Table~\ref{tab:validation-results})


\begin{table*}[t]
  \caption{Assembly Feasibility of AI-Generated Objects Without Individual Geometric Processing Modules}
  \label{tab:validation-results}
  \centering
  \begin{tabular}{l c c c c c}
    \toprule
    Object & \shortstack{Without\\Component Count Check} & \shortstack{Without\\Overhang Detection} & \shortstack{Without\\Vertical Stack Check} & \shortstack{Without\\Connectivity Search} & \shortstack{Complete Assembly \\ Pipeline}\\
    \midrule
    Stool    & Failed & Passed & Passed & Failed & Passed \\
    Shelf    & Failed & Failed & Passed & Failed & Passed \\
    Letter T & Passed & Passed & Failed & Passed & Passed \\
    Table    & Passed & Passed & Passed & Failed & Passed \\
    \bottomrule
  \end{tabular}
\end{table*}

\subsubsection{Component Count}
To ensure that AI-generated designs are feasible with a limited set of physical parts, we implement a component count check that compares the number of required components against a pre-defined inventory limit. As shown in Table~\ref{tab:ablation-component}, the \textit{stool} and \textit{shelf} exceeded the 40-component inventory limit and failed the component count check. Without this check, the assembly would stop due to insufficient physical parts. With the check enabled, the system triggers automatic geometric rescaling to reduce the component count to within allowable limits (e.g., the \textit{stool} was scaled from 45 to 30 \textit{components} , and the shelf was scaled from 60 to 32 components). The \textit{letter T} and \textit{table} remained within the limits and passed without modification. These results highlight the importance of component count validation in adapting generative outputs to physical constraints.

\begin{table} [ht]
  \caption{Component Count Check and Rescaling}
  \label{tab:ablation-component}
  \centering
  \begin{tabular}{l c c c}
    \toprule
    Object & \shortstack{Baseline\\Assembly }& \shortstack{Component\\Count Check} & \shortstack{Component\\Count Rescaling} \\
    \midrule
    Stool    & 45 Components& Fail & 30 Components\\
    Shelf    & 60 Components& Fail & 32 Components\\
    Letter T & 8 Components& Pass & 8 Components\\
    Table    & 21 Components& Pass & 21 Components\\
    \bottomrule
  \end{tabular}
\end{table}

\subsubsection{Overhang Detection}
To prevent structural instability during assembly, we apply an overhang check that limits unsupported cantilevers to a maximum span of three units. Table~\ref{tab:ablation-overhang} shows that only the \textit{shelf} failed the overhang check. Its horizontal spans exceeded the 3-unit cantilever limit identified through physical testing. The system responded by rescaling along the overhanging axis. Other objects passed without modification. The overhang detection is necessary for geometries with wide horizontal features.
\begin{table}[h]
  \caption{Overhang Check and Rescaling}
  \label{tab:ablation-overhang}
  \centering
  \begin{tabular}{l c c c}
    \toprule
    Object & \shortstack{Baseline\\Assembly}& \shortstack{Overhang\\Check} & \shortstack{Overhang\\Rescaling} \\
    \midrule
    Stool    & 45 Components
& Pass & 45 Components\\
    Shelf    & 60 Components& Fail & 54 Components\\
    Letter T & 8 Components
& Pass & 8 Components\\
    Table    & 21 Components& Pass & 21 Components\\
    \bottomrule
  \end{tabular}
\end{table}

\subsubsection{Vertical Stack Detection}
The vertical stack constraint limits unsupported columns to four components. As shown in Table~\ref{tab:ablation-vertical}, the \textit{letter T} and \textit{table} failed, each containing unstable 5-high stacks. After rescaling the geometry along the z axis, the number of components in the new assemblies was reduced to 7 and 20, respectively. The \textit{stool} and \textit{shelf} passed unchanged. This module prevents instability in assemblies with tall, narrow elements.


\begin{table}[h]
  \caption{Vertical Stack Check and Rescaling}
  \label{tab:ablation-vertical}
  \centering
  \begin{tabular}{l c c c}
    \toprule
    Object & \shortstack{Baseline\\Assembly} & \shortstack{Vertical Stack\\Check} & \shortstack{Vertical Stack\\  Rescaling}\\
    \midrule
    Stool    & 46 Components
& Pass & 46 Components\\
    Shelf    & 60 Components
& Pass & 60 Components\\
    Letter T & 8 Components
& Fail & 7 Components\\
    Table    & 21 Components& Fail & 20 Components\\
    \bottomrule
  \end{tabular}
\end{table}

\subsubsection{Connectivity-Aware Assembly Sequencing}
To ensure structural integrity and reachability during construction, the system employs a connectivity-aware sequencing strategy that prioritizes grounded and accessible components.  When using a simple sorting strategy (placing components in increasing order of z, then x, then y), some assemblies fail since components might be placed mid-air. For example, in letter T, naive sorting causes the robot to place the outer ends of the horizontal bar before the center, which results in disconnected components since the outer parts are not yet structurally connected to the vertical column. In contrast, connectivity-aware sequencing prioritizes the central segment closest to the column, ensuring that all parts remain structurally grounded as the assembly progresses. Similar issues occur in the \textit{stool} and \textit{table}, where naive ordering leads to placements of components that aren't connected. Only the \textit{shelf} succeeds under naive sorting, because its geometry aligns with the sort direction and has fewer elevated or branching components. When connectivity-based sequencing is enabled, all assemblies are completed successfully. This demonstrates that assembly depends on proper sequencing, particularly in branching or overhanging assemblies.

These fabrication constraints may seem like small concerns, but they are crucial for AI-generated designs, as AI does not inherently account for them. Although a human can manually modify a geometry if something does not work, it is essential to develop automated failure handling approaches that can automatically adjust the assembly geometry to ensure assembly feasibility for AI-generated objects. Currently, we use an algorithmic approach. Future studies could explore the integration of this process with a physics-based simulation environment.

\subsubsection{Calibrating Robot Motion for On-Demand Assembly}

\begin{figure}[b]
    \centering
    \includegraphics[width=1\linewidth]{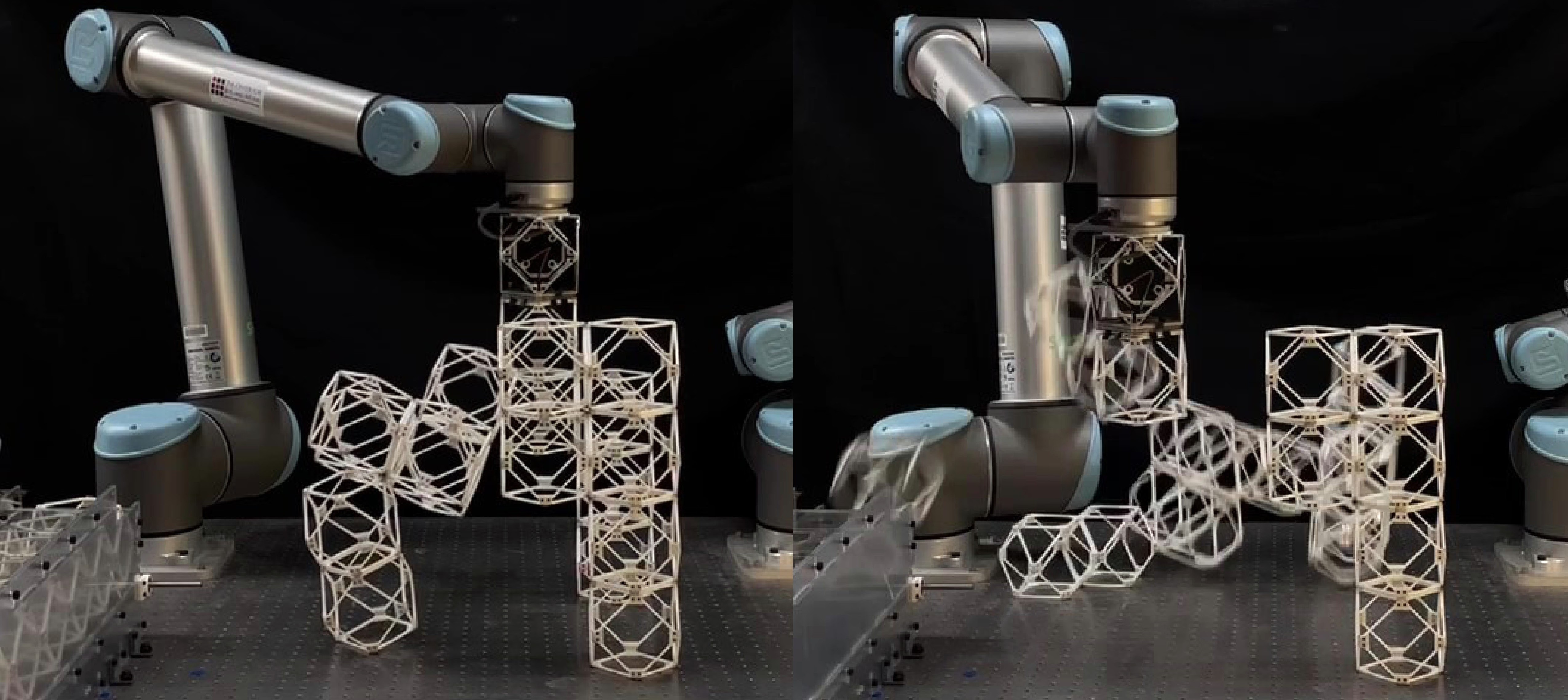}
    \caption{Assembly failing at different stages due to impact and vibration during calibration }
    \Description{Photo showing assembly failing at different stages due to impact and vibration during calibration.}
    \label{fig:cali}
\end{figure}

Calibration of the robotic arm's speed and acceleration is important to avoid failure. While increasing speed reduced overall assembly time, it also introduced instability, particularly when placing cantilevered components (Fig. \ref{fig:cali}). Assembly failures can result from vibration or impact forces caused by both the robot arm and the components. Although various control strategies can improve stability and precision, this paper shows that a basic trial-and-error speed calibration meets the minimum requirement for successful assembly.

We completed a basic calibration using an AI-generated stool as the test object. The process begins with an initial velocity of 1 mm/s, gradually increasing in increments of 0.5 mm/s until a failure is detected. Acceleration is calibrated at fixed ratios of 1:1 or 1:2 relative to the maximum velocity. The calibration results are presented in Table \ref{tab:velocity_acceleration}. Failures occurred when the max velocity is at 2.0 mm/s with a velocity-to-acceleration ratio of 1:1, and at 2.5 mm/s with a velocity-to-acceleration of 2:1. Based on these results, a maximum velocity of 2 mm/s and an acceleration of 1 mm/s\textsuperscript{2} were used to successfully assemble the objects shown in  (Fig. \ref{fig:all}). 

\begin{table}[h]
  \caption{Velocity, Acceleration, and Assembly Failure}
  \label{tab:velocity_acceleration}
  \centering
  \begin{tabular}{c c c c c}
    \toprule
    \multicolumn{3}{c}{\shortstack{Velocity to Acceleration\\Ratio 1:1}} & \multicolumn{2}{c}{\shortstack{Velocity to Acceleration\\Ratio 2:1}} \\
    \cmidrule(lr){1-3} \cmidrule(lr){4-5}
    Velocity & Acceleration & Assembly & Acceleration & Assembly \\
    \midrule
    1.0 mm/s & 1.0 mm/s$^{2}$ & Pass & 0.5 mm/s$^{2}$  & Pass \\
    1.5 mm/s & 1.5 mm/s$^{2}$ & Pass & 0.75 mm/s$^{2}$ & Pass \\
    2.0 mm/s & 2.0 mm/s$^{2}$ & Fail & 1.0 mm/s$^{2}$  & Pass \\
    2.5 mm/s & 2.5 mm/s$^{2}$ & Fail & 1.25 mm/s$^{2}$ & Fail \\
    \bottomrule
  \end{tabular}
\end{table}

Although calibrating the robot’s speed may appear trivial, we see it as an essential step to ensure reliable interactions between the robot and the materials, especially in an automated pipeline like Speech to Reality. In this work, we demonstrate that a simple trial-and-error calibration is effective to determining a workable velocity and acceleration. Because our components use passive magnetic alignment and tolerate small pose errors, we found that a simple empirical tuning of velocity/acceleration was sufficient. However, future studies could explore more advanced control strategies, such as Model Predictive Control or motion/force hybrid control, if higher precision or adaptability becomes necessary.

\subsection{On-Demand Assembly and Speed}

\begin{figure}[t]
    \centering
    \includegraphics[width=1\linewidth]{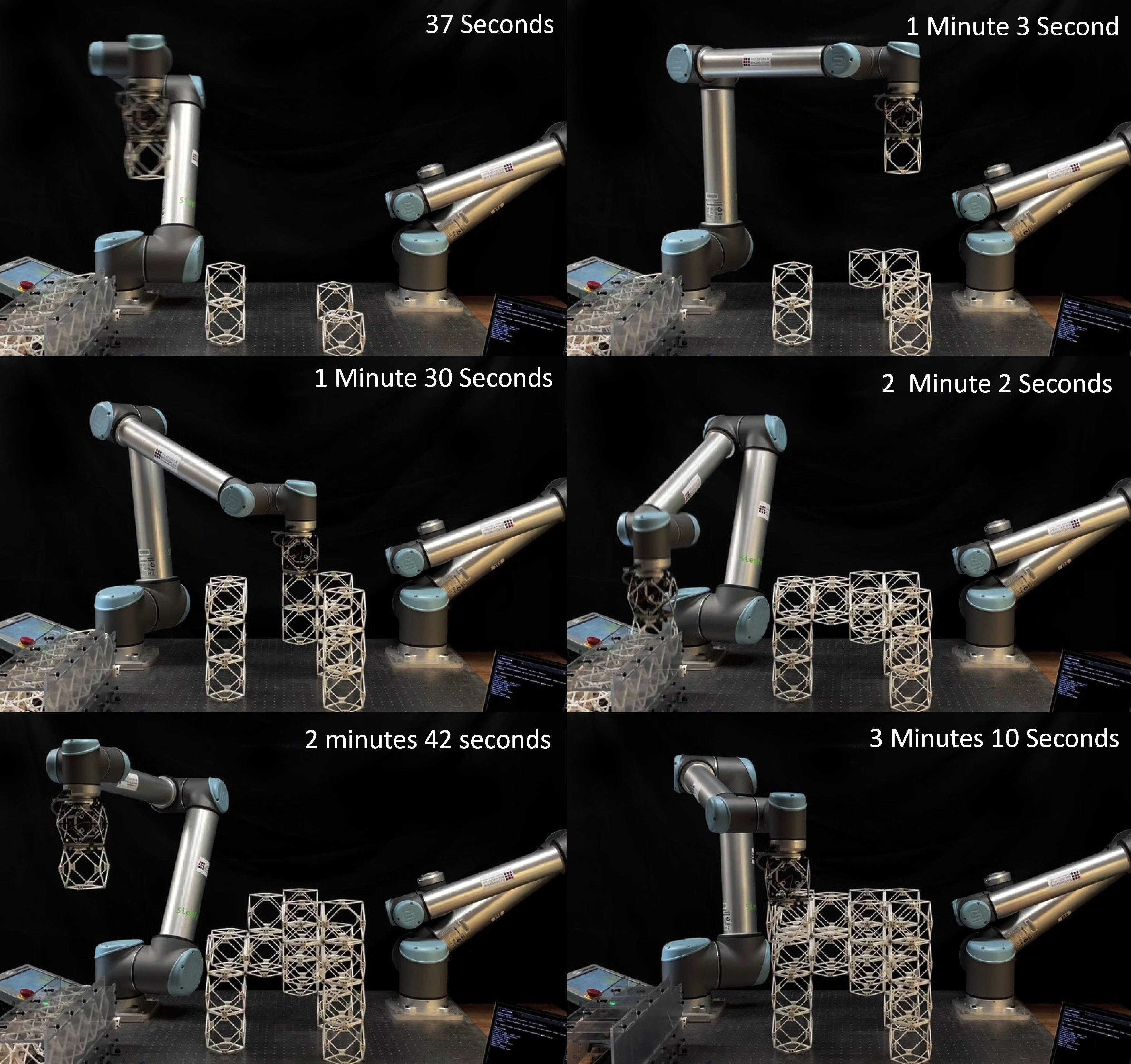}
    \caption{Assembly of “A Simple Stool” at different stages.
Total Assembly Time: 3 Minutes and 36 Seconds}
    \Description{Assembly of “A Simple Stool” at different stages.}
    \label{fig:stool}
\end{figure}

\begin{figure} [!]
    \centering
    \includegraphics[width=1\linewidth]{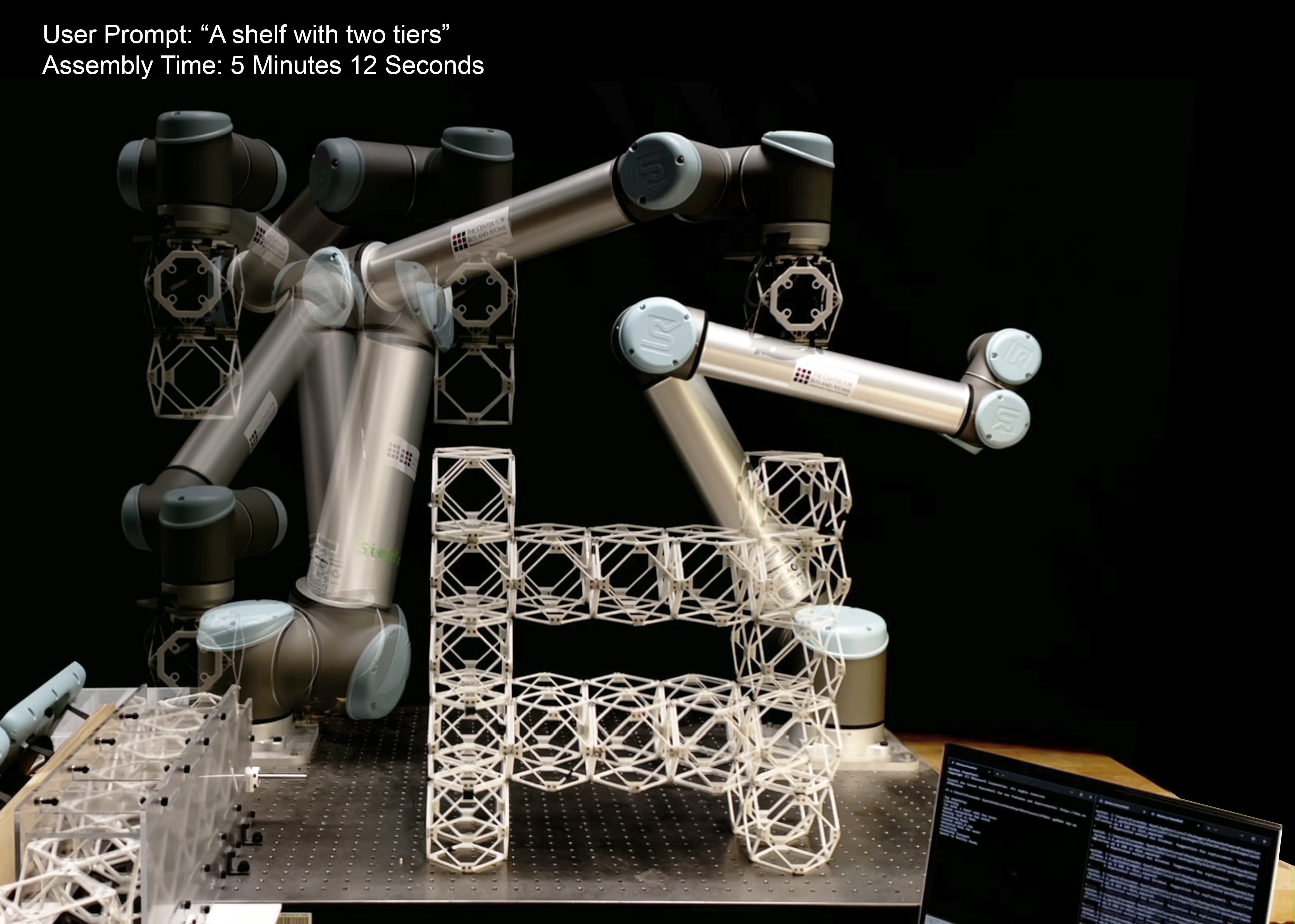}
    \caption{User Prompt "Assemble me a shelf with two tiers”. Assembly Time: 5 minute 12 seconds}
    \Description{Shelf Assembly Process}
    \label{fig:shelf}
\end{figure}

\begin{figure}[!]
    \centering
    \includegraphics[width=1\linewidth]{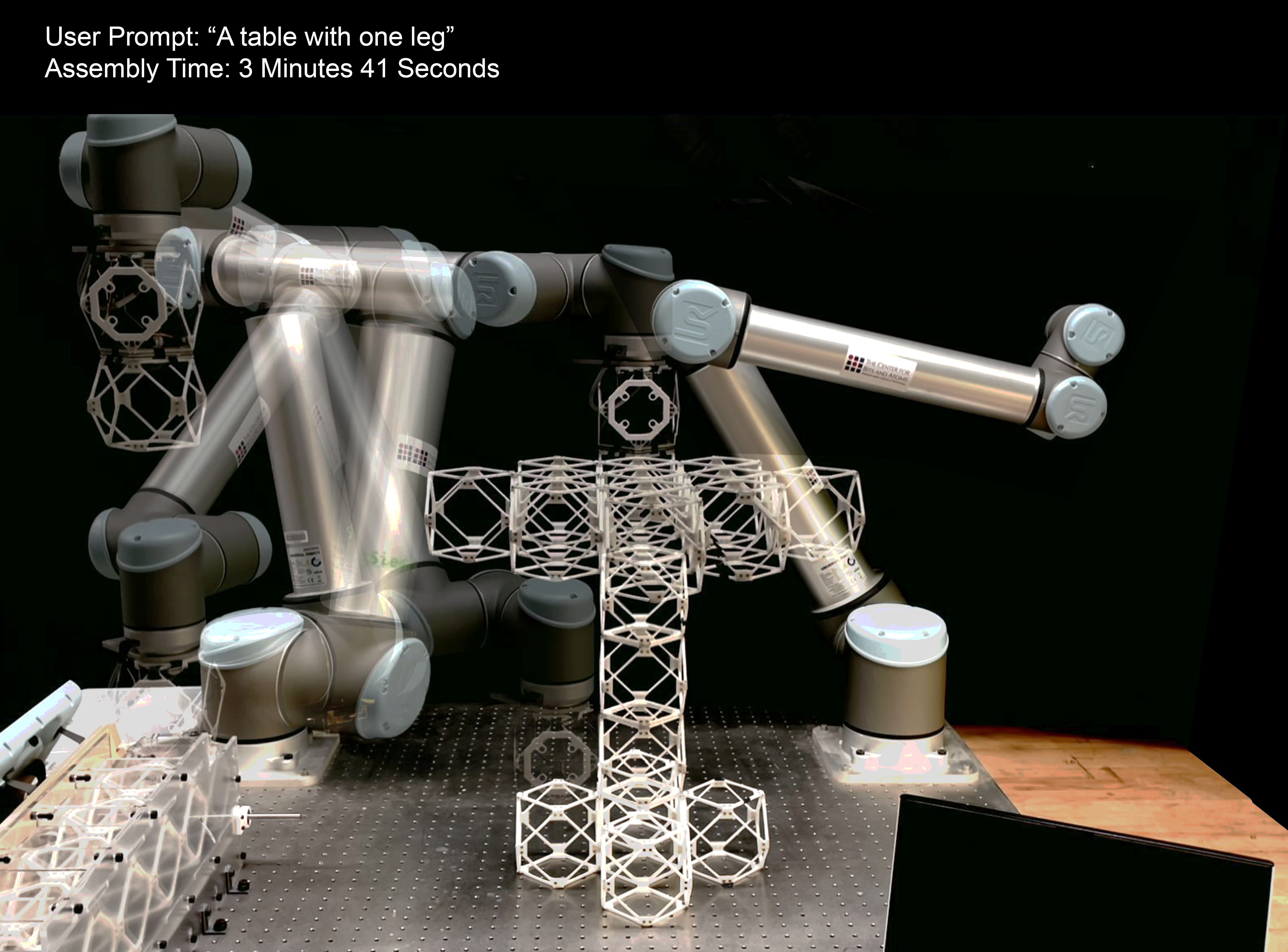}
    \caption{User Prompt "Assemble me a table with one leg”. Assembly Time: 3 minute 41 seconds}
    \Description{A photo of a table with one leg, assembled in 3 minutes and 41 seconds.}
    \label{fig:tablee}
\end{figure}

\begin{figure} [!]
    \centering
    \includegraphics[width=1\linewidth]{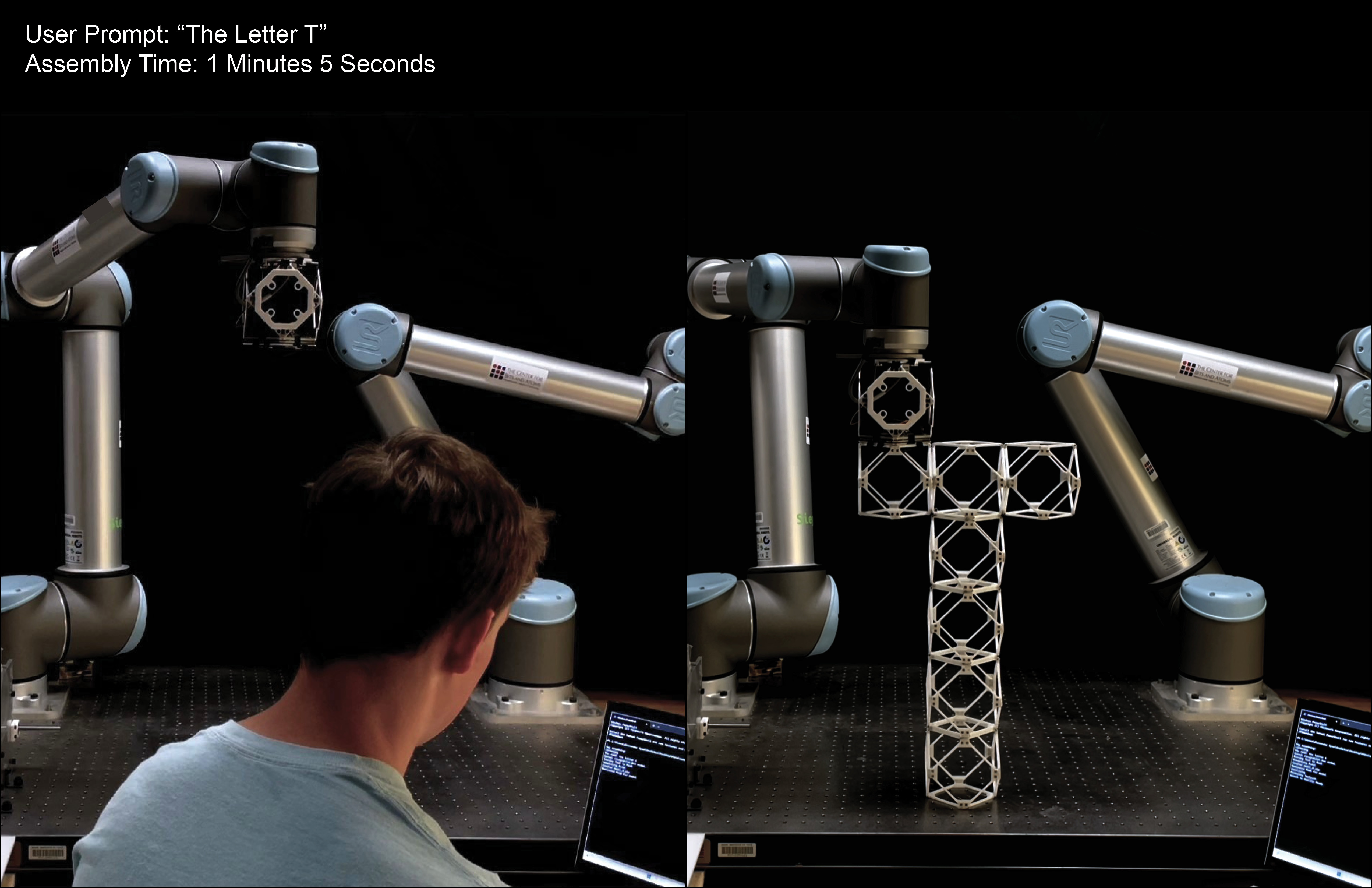}
    \caption{User Prompt: The Letter T. Assembly Time: 1 Minute 5 Seconds}
    \Description{A photo of the letter T, assembled in 1 minute and 5 seconds.}
    \label{fig:t}
\end{figure}

While 3D generative AI can rapidly create 3D models, making physical objects can be time-consuming. Discrete robotic assembly could potentially complement the rapid generation capacity of AI by reducing production time by composing larger objects from smaller pre-fabricated components. In Table~\ref{tab:assembly-time-comparison}, we demonstrate that discrete assembly can construct objects with volumes ranging from 4279 cm3 to 7356 cm3 in 1 to 5 minutes  (Fig. \ref{fig:stool})  (Fig. \ref{fig:shelf}) (Fig. \ref{fig:tablee}) (Fig. \ref{fig:t}). Although discrete robotic assembly requires prior fabrication of the modular components, we note that other methods, such as 3D printing, also involve significant preparation time (e.g., filament loading, bed leveling, resin curing). To ensure a fair comparison and user-facing experience, we measure fabrication time only from the moment a user requests an object.

\begin{table}[H]
  \caption{Discrete Robotic Assembly Time for AI-Generated for from Various User Prompts}
  \label{tab:assembly-time-comparison}
  \centering
  \begin{tabular}{p{0.5\columnwidth}c c}
    \toprule
    User Prompt & Object Volume & Time \\
    \midrule
    I want a simple stool & 6496 cm\textsuperscript{3} & 3m 36s\\
    A shelf with two tiers & 11831 cm\textsuperscript{3} & 5m 12s \\
    The letter ‘T’ & 4279 cm\textsuperscript{3} & 1m 05s \\
    Assemble me a table with one leg & 7356 cm\textsuperscript{3} & 3m 41s \\
    \bottomrule
  \end{tabular}
\end{table}


Currently, the resolution and speed of discrete robotic assembly is controlled by component size. While the current component size works for demonstrating the objects in (Fig. \ref{fig:all}), follow-up studies can investigate the use of smaller components to improve the resolution. 
Future studies can also provide comparative insight into the trade-offs between resolution, speed, and efficiency. 

Direct comparison to other fabrication processes, such as 3D printing, is difficult due to differences in resolution and the material properties of the final output. As a reference point, simulating a print of the stool on a large format FDM printer in standard settings produced an estimated build time of 3~days 1~hour 45~minutes. Printed parts would have finer resolution and different mechanical behavior than discretely assembled structures.

This paper does not aim to replace traditional manufacturing. Instead, it explores the potential of an automated pipeline to translate AI-generated outputs into physical forms. The results mark a step toward integrating 3D generative AI with discrete robotic assembly to support rapid on-demand production. The on-demand nature of the project enables faster feedback loop between humans and AI for collaborative co-creation. 


\subsection{Modularity, Reuse, and Sustainable Production of AI-Generated Objects  }

Every object created through the speech-to-reality system was assembled using the same set of 40 reusable components \ref{fig:all}. The re-usability of components demonstrates the potential to scale production in line with the output capacity of Generative AI without increasing material waste. Through these demonstrations, we successfully performed non-destructive assembly across multiple objects, confirming that our components and robotic end effector can be used more than once. 

Components are reused for each assembly by disassembling the object and placing them on the conveyor belt for the next build. The conveyor belt system played a crucial role in enabling efficient material handling and reuse. While we manually disassembled objects in this study, future research can explore using the robotic arm for disassembly or modifying an existing assembly with generative AI and speech commands. 

\section{Limitations and Future Work}

Although the speech-to-reality system demonstrates initial feasibility, it has several limitations that can guide future research. 

The resolution and expressiveness of the fabricated output are constrained by the size and geometry of the modular components. Although this approach could enable on-demand assembly and rapid prototyping of large objects, it is limited in fidelity. The objects might also be less suitable for applications requiring precise tolerances or specialized material properties, which techniques like 3D printing or machining are better equipped to provide. Future research might explore hybrid workflows that combine discrete assembly with additive or subtractive techniques, enabling finer details where necessary.

Robotic control strategies also present a potential opportunity for improvement. In the current implementation, simple velocity and acceleration tuning was sufficient due to the predictability of the components. However, more fragile or irregular assembly components will require better motion planning, dynamic adaptation, and feedback control. Methods such as learning-based control or force-aware assembly strategies could expand the applicability of the system to the assembly of more delicate materials and components in the future. 

Although our pipeline supports object reuse for circularity, it currently relies on manual disassembly. Fully closing the loop where robotic agents not only assemble, but also disassemble or modify existing structures could open up possibilities for dynamic environments where AI adapts and reshapes objects over time in response to user needs. This could also enable more dynamic, bidirectional workflows in which AI, robots, and humans collaboratively iterate on physical designs over time. Rather than treating object creation as a fixed endpoint, such a system could support continuous refinement through natural language commands and environmental feedback. Incorporating reversible assemblies and interactive editing would further strengthen the system’s potential for sustainable, adaptive, and creative human–machine collaboration. 

In addition to speech input, other modalities such as gestures or visual cues can be integrated to enable greater human control and better capture human intent \cite{kyaw_gesture_2023}. This multimodal interaction could allow users to more precisely specify geometry and spatial relationships, making the system more intuitive and responsive. Additionally, the current system follows a one-way trajectory from input to output, limiting opportunities for preview and iteration. Interfaces such as augmented reality could help bridge this gap by allowing users to visualize, simulate, and refine structures before physical assembly begins \cite{kyaw_humanmachine_2024}. 

Finally, comparing this approach with established fabrication methods like 3D printing is inherently complex. Differences in material, resolution, and functional outcomes make direct benchmarks difficult. Future work could systematically explore these comparative dimensions through controlled user studies and task-specific evaluations. Overall, the results presented here mark an early but promising work toward systems that connect generative AI with the physical world as a framework for rapid, iterative, sustainable, and collaborative creation between humans, AI, and machines.

\section{Conclusion}

This paper introduces Speech-to-Reality, an automated system that transforms spoken object requests into modular physical assemblies by connecting 3D generative AI and discrete robotic assembly. By translating natural language into tangible output, the system bridges the gap between AI-driven design and on-demand physical production. In summary, this paper presents three main outcomes. 
\begin{itemize}
  \item We present an end-to-end pipeline that integrates natural language input, 3D generative AI, component discretization, geometric processing, path planning, and robotic assembly. Hardware integration includes reusable voxel components, a custom robotic end effector, and a conveyor system. 
  \item We identify key fabrication constraints and develop geometric processing methods to enable a feasible robotic assembly of AI-generated meshes. The system automatically modifies AI-generated assemblies to account for real-world constraints such as inventory limits, overhang stability, vertical stacking, connectivity, and robotic reachability.
  \item We evaluated different geometric processing methods, user prompts, and assembly time. The results show that the system can assemble a range of objects within minutes, demonstrating its potential for on-demand, sustainable production aligned with the generative capabilities of AI.
\end{itemize}

The research serves as a framework and outlines the essential steps for integrating AI-driven generative design with robotic fabrication. This work points towards a future of AI-driven on-demand robotic fabrication and offers a different perspective on bridging the gap between digital design and physical realization. 

\begin{acks}
This research is supported by CBA Consortia funding and the MIT Morningside Academy of Design. 
\end{acks}

\bibliographystyle{ACM-Reference-Format}
\bibliography{references2}

\end{document}